\RequirePackage{amsthm}
\documentclass[pdflatex,iicol,sn-mathphys]{sn-jnl}

\usepackage{graphicx}%
\usepackage{multirow}%
\usepackage{amsmath,amssymb,amsfonts}%
\usepackage{amsthm}%
\usepackage{mathrsfs}%
\usepackage[title]{appendix}%
\usepackage{xcolor}%
\usepackage{textcomp}%
\usepackage{manyfoot}%
\usepackage{booktabs}%
\usepackage{algorithm}%
\usepackage{algorithmicx}%
\usepackage{algpseudocode}%
\usepackage{listings}%
\usepackage{amsmath,bm}

\theoremstyle{thmstyleone}%
\theoremstyle{thmstyletwo}%

\theoremstyle{thmstylethree}%

\begin{document}

\title[Article Title]{CogCartoon: Towards Practical Story Visualization}


\author*[1]{\fnm{Zhongyang} \sur{Zhu}}\email{zhuzy.2020@tsinghua.org.cn}

\author*[2]{\fnm{Jie} \sur{Tang}}\email{jietang@tsinghua.edu.cn}

\affil[1]{\orgname{Zhipu AI}, \orgaddress{\city{Beijing}, \postcode{100084}, \country{China}}}

\affil[2]{\orgdiv{Department of Computer Science and Technology}, \orgname{Tsinghua University}, \orgaddress{\city{Beijing}, \postcode{100084}, \country{China}}}

\abstract{The state-of-the-art methods for story visualization demonstrate a significant demand for training data and storage, as well as limited flexibility in story presentation, thereby rendering them impractical for real-world applications. We introduce \textbf{CogCartoon}, a practical story visualization method based on pre-trained diffusion models. To alleviate dependence on data and storage, we propose an innovative strategy of character-plugin generation that can represent a specific character as a compact 316 KB plugin by using a few training samples. To facilitate enhanced flexibility, we employ a strategy of plugin-guided and layout-guided inference, enabling users to seamlessly incorporate new characters and custom layouts into the generated image results at their convenience. We have conducted comprehensive qualitative and quantitative studies, providing compelling evidence for the superiority of CogCartoon over existing methodologies. Moreover, CogCartoon demonstrates its power in tackling challenging tasks, including long story visualization and realistic style story visualization.}

\keywords{story visualization, pre-trained diffusion models, character-plugin generation, plugin-guided and layout-guided inference}



\maketitle

\section{Introduction}\label{sec1}

Recently, the advancement of story visualization has been significantly propelled by the utilization of large-scale pretrained transformers \citep{cc2,cc1,cc3} and diffusion models \citep{cc4,cc5,cc6,cc7,cc8,cc9,cc10}, exemplified by StoryDALL-E \citep{cc11}, Make-A-Story \citep{cc12} and ARLDM \citep{cc13}. These methodologies demonstrate the capability to generate a coherent sequence of visuals through training pre-existing text-to-image models on a specific dataset. However, they are impractical in real-world scenarios owing to the following two drawbacks.

First, the success of these approaches heavily relies on the availability of extensive training data and ample storage resources. Regarding the aspect of training data, two widely used datasets, namely FlintstonesSV \citep{cc14} with 20132 training samples and PorotoSV \citep{cc15} with 10191 training samples, are commonly employed. However, the collection of tens of thousands of samples is not feasible at the initial stage of a story book creation. As for the aspect of storage resources, the implementation of these methods involves storing a distinct model for each individual story, making them unsuitable for large-scale commercial scenarios where there are often numerous independent stories.

Second, these approaches demonstrate limited flexibility in terms of incorporating new characters and controlling layout. In practical applications, the demand for plugging in new characters and controlling layouts anytime and anywhere is highly pronounced among users. Unfortunately, these approaches face challenges in meeting the aforementioned requirements due to their reliance on fine-tuning pre-trained models with character-specific datasets and their failure to incorporate layout control strategies.

\begin{figure*}[ht]%
\centering
\includegraphics[width=0.99\textwidth]{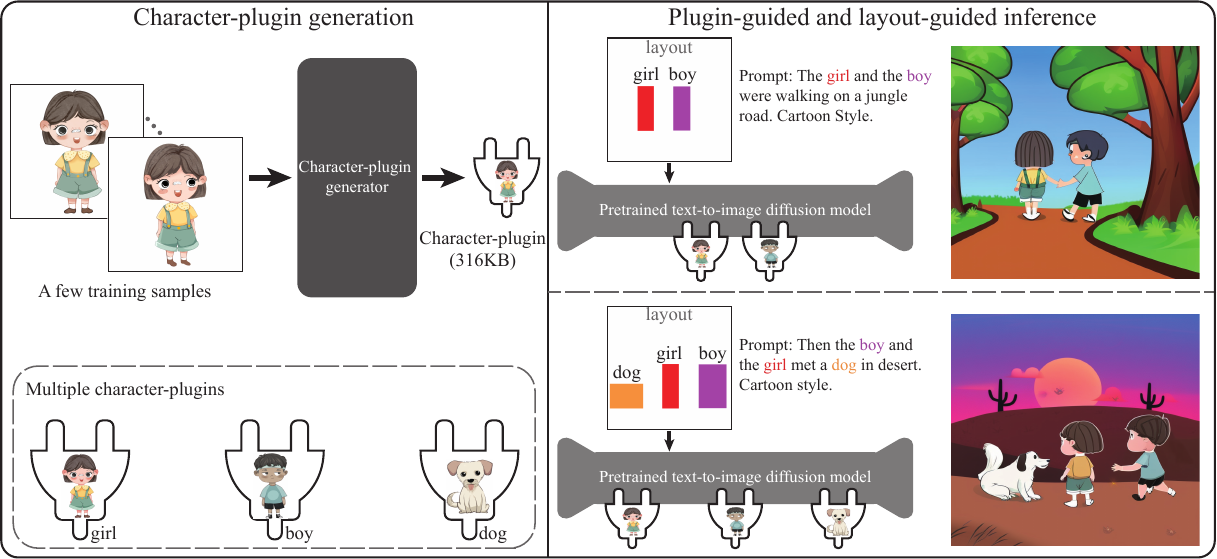}
\caption{Demonstration of the CogCartoon.}\label{fig1}
\end{figure*}

In this study, we propose an innovative and practical story visualization framework named \textbf{CogCartoon} to simultaneously address the aforementioned limitations. As depicted in Fig. \ref{fig1}, CogCartoon incorporates two innovative strategies, namely character-plugin generation and plugin-guided and layout-guided inference. In character-plugin generation, the representation of a specific character can be achieved through a compact 316 KB plugin, utilizing only a limited number of training samples. The storage for multiple independent stories solely involves the inclusion of various character-plugins and a shared diffusion model. Hence, the reliance on data and storage is mitigated. In plugin-guided and layout-guided inference, users are granted the flexibility to add new characters and modify the positions of characters at their convenience. Specifically, when introducing a new character into the story, users can effortlessly create a corresponding plugin by providing a small number of samples. Subsequently, employing the proposed inference method enables simultaneous utilization of the newly created character-plugin, existing character-plugins, and custom layout to generate an illustrative depiction of the story with the newly introduced character seamlessly integrated within the preset layout.

Due to the insufficient size of training datasets in experiment, state-of-the-art story visualization methods, such as StoryDALL-E \citep{cc11} and Make-A-Story \citep{cc12}, cannot be served as comparative baselines. Recent work \citep{cc20} claimed that customized image generation methods \citep{cc16,cc17,cc18,cc19} could be served as potential solutions for visualizing story under the condition of limited training samples. Consequently, the most advanced techniques for customized image generation are chosen as baselines, yet experimental result reveals their inapplicability within the domain of story visualization. Furthermore, we also evaluate CogCartoon on challenging tasks including long story visualization and realistic style story visualization, and the experimental results demonstrate its superior performance.

In conclusion, we offer the following contributions in this paper:

\begin{enumerate}[1.]
\item We systematically analyze the drawbacks of existing story visualization methods in terms of practicability and provide detailed explanations for their limitations in training data, storage, and flexibility, aiming to pave the way for a practicality-centric story visualization method.
\item We propose CogCartoon, a practical method for visualizing stories. It relies on a limited amount of training data and storage and allows users greater freedom in visualizing their stories.
\item We demonstrate that CogCartoon surpasses the latest customized image generation methods, such as Custom Diffusion \citep{cc18} and Cones2 \citep{cc19}, in terms of both qualitative and quantitative analyses.
\item We set up two challenging tasks, including long story visualization and realistic style story visualization. Experimental results demonstrate the superior performance of CogCartoon in both tasks.
\end{enumerate}

\section{Related Works}\label{sec2}

\subsection{Text-to-Image Generation}\label{subsec21}

The goal of text-to-image generation is to produce a visually compelling images based on a given textual input \citep{cc2}. The initial investigation into text-to-image generation predominantly utilize GAN-based methodologies \citep{cc21,cc22,cc23,cc24,cc25,cc26,cc27,cc28}. However, the variety of images produced with the GAN-based methods is limited because it is difficult for the adversarial learning mechanism to converge on large-scale and complex datasets \citep{cc10}. Diffusion method \citep{cc4} effectively addresses this issue and facilitates the widespread commercial utilization of text-to-image generation \citep{cc7,cc9,cc10,cc29}. Exemplary works, such as Imagen \citep{cc29}, Glide \citep{cc7}, DALL.E 2 \citep{cc9} and Stable Diffusion \citep{cc10}, have the capability to generate a single visual representation according to a given prompt. However, they are unable to create a sequence of coherent pictures based on a description of story.

\subsection{Coherent Story Visualization}\label{subsec22}

Earlier studies \citep{cc30,cc31,cc32,cc33,cc34,cc35} have focused on enhancing the GAN networks to produce visually coherent images. However, the story generation methods based on GAN networks fail to produce satisfactory results due to the inherent limitations of GAN architecture. To improve the story visualization performance, STORYDALL-E \citep{cc11} employs a more powerful foundation model named DALL-E \citep{cc1}. After that, the proposal and development of diffusion models \citep{cc4,cc5,cc6,cc7,cc8,cc9,cc10} have propelled the advancement of story visualization \citep{cc12,cc13,cc36,cc37}. Make-A-Story \citep{cc12} and ARLDM \citep{cc13} employ an auto-regressive way to train a pre-existing diffusion model. These methods empower the pre-trained diffusion model to generate coherent images by leveraging historical outcomes. However, as previously mentioned, these methods necessitate substantial data and storage while lacking flexibility in accommodating new characters and controlling layout. Consequently, their practical implementation becomes challenging. One concurrent study \citep{cc20} endeavors to generate coherent images under limited number of training samples. This approach employs a two-stage training process, encompassing component training and character training, as well as a two-stage inference procedure involving text-to-image generation and inpainting. Moreover, the approach is limited to visualize the narratives of merely two characters. CogCartoon features concise one-stage training and inference processes, showcasing its capacity to visually depict narratives involving up to three characters.

\begin{figure*}[ht]%
\centering
\includegraphics[width=0.99\textwidth]{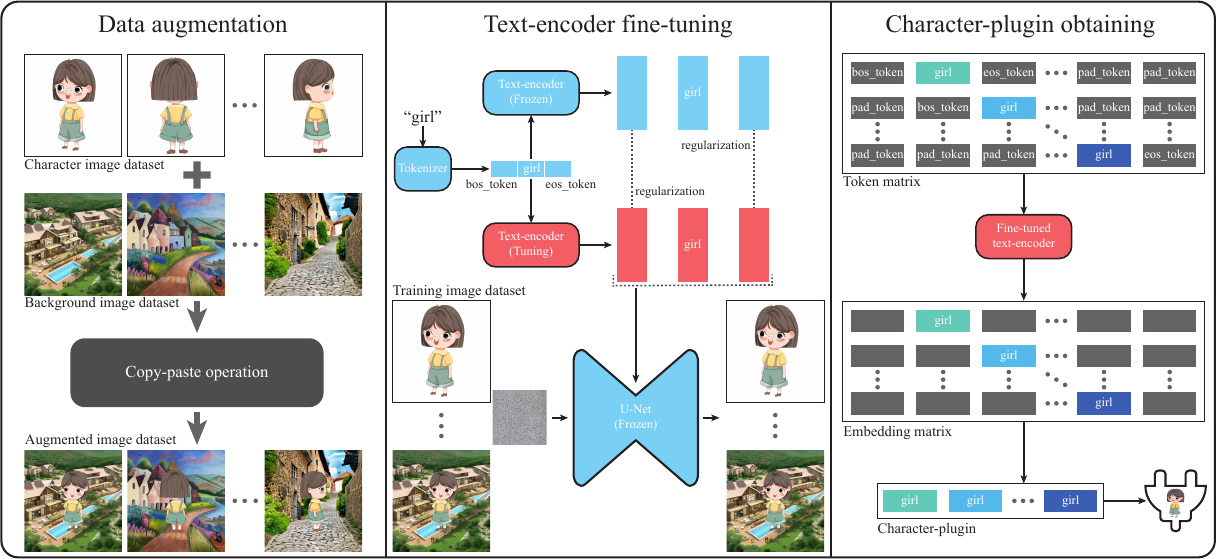}
\caption{Demonstration of the strategy of character-plugin generation.}\label{fig2}
\end{figure*}

\subsection{Customized Image Generation}\label{subsec23}

By utilizing a limited amount of training data, customized image generation techniques \citep{cc18,cc19} can proficiently produce images containing user-specified subjects. The introduction of Dreambooth \citep{cc38} and Text Inversion \citep{cc39} signifies the inception of early endeavors in customizing diffusion models. These methods exhibit strong performance in single-subject customization tasks, but demonstrate limited efficacy in multi-subject customization tasks. To enhance the performance in multi-subject customization, Custom Diffusion \citep{cc18} focuses on optimizing a limited number of parameters within the diffusion model and proposes a novel approach that combines multiple fine-tuned models. Cones 2 \citep{cc19} employs the residual token embedding to represent subjects and utilizes a layout guidance method for effectively presenting multiple subjects in the generated image. Customized image generation methods, due to their minimal data requirements and the capability to incorporate subjects into the generated results, hold potential for visualizing narratives \citep{cc20}. However, our experimental findings indicate that the advanced customized image generation methods are not applicable within the domain of story visualization.

\section{Method}\label{sec3}

In this section, we provide the details of CogCartoon. CogCartoon aims to generate coherent images that contain any combination of a given set of characters, leveraging limited training samples provided for each character. The demonstration of CogCartoon is illustrated in Fig. \ref{fig1}. CogCartoon consists of two strategies: character-plugin generation and plugin-guided and layout-guided inference, which are introduced in detail in Sections \ref{subsec31} and \ref{subsec32}, respectively.

\subsection{Character-Plugin Generation}\label{subsec31}

Character-plugin generation is capable of characterizing a character as a lightweight plugin with just a few training samples. The strategy of character-plugin generation is demonstrated in Fig. \ref{fig2}. It consists of three core steps: data augmentation, text-encoder fine-tuning, and character-plugin obtaining. Next, we will provide the details for each step.

\subsubsection{Data Augmentation}\label{subsec311}

The copy-paste operation is a straightforward and efficient data augmentation technique that finds extensive application in computer vision tasks \citep{cc40,cc41,cc42,cc43}. We innovatively employ this technique in the character-plugin generation. The left part of the Fig. \ref{fig2} depicts the flow of the data augmentation. 

\textbf{\emph{First}}, we use ChatGLM \citep{cc44} and Stable Diffusion \citep{cc10} to generate a background image dataset. Specifically, ChatGLM is initially employed to generate a substantial number of sentences that describe diverse natural scenes, followed by the utilization of Stable Diffusion to produce corresponding background images based on the scene descriptions. 

\textbf{\emph{Second}}, we employ a copy-paste operation to generate an augmented image dataset. We employ a random selection process to choose one image from the background image dataset and another from the character image dataset. Then, we incorporate the character into the central region of the background image, resulting in an augmented image. The rationale behind placing the character in the central position of the background is rooted in previous research findings, which have demonstrated that determining object positioning within the background is unnecessary \citep{cc40}. By employing a sample copy-paste operation, algorithmic efficacy can be enhanced without considering object placement.

\textbf{\emph{Third}}, we combine the character image dataset $D_{char}^m$ with the augmented image dataset $D_{aug}^n$ to yield a training image dataset $D_{train}^q$:

\begin{equation}
D_{train}^q=D_{char}^m\cup D_{aug}^n,\label{eq1}
\end{equation}
where $m$, $n$ and $q$ indicate the number of samples in character image dataset, augmented image dataset and training image dataset, respectively.

\subsubsection{Text-Encoder Fine-Tuning}\label{subsec312}

The central portion of Fig. \ref{fig2} illustrates the conceptual framework of the text-encoder fine-tuning. We initially assign a [class noun] label to all images in training dataset, where [class noun] represents a high-level class descriptor of the character (e.g., girl, boy, dog, etc.). Then, we train the text-encoder with the labeled training dataset based on a subject-preservation loss $L_{sub}$ and a regularization loss $L_{reg}$ \citep{cc19}. The two losses are respectively defined as:
\begin{equation}
L_{sub}\left(E^t\right)=\mathbb{E}_{\left(x,c,\epsilon,t\right)}[\|\ \epsilon_\theta(x_t,E^t(c),t)-\epsilon\|_2^2],\label{eq2}
\end{equation}
and
\begin{equation}
L_{reg}\left(E^t\right)=\mathbb{E}_{\left(c\right)}[\sum_{c\in N C T}{\|E^t(c)-E^f(c)\|_2^2}],\label{eq3}
\end{equation}
where $E^t$ represents the text-encoder that requires fine-tuning; $x\in D_{train}^q$; $c="class\ noun"$; $\epsilon\in\mathcal{N}(0,1)$; $NCT$ refers to non-character tokens such as \verb|bos_token| and \verb|eos_token|; $E^f$ indicates the frozen text-encoder in original Stable Diffusion pipeline. Similar to Cones 2 \citep{cc19}, the total loss can be given as:
\begin{equation}
L_{total}=L_{sub}+\lambda L_{reg},\label{eq4}
\end{equation}
where the parameter $\lambda$ is utilized to regulate the magnitude of the regularization loss. 

The Stable Diffusion pipeline, equipped with the fine-tuned text-encoder after the process of fine-tuning, is capable of generating images that depict a specific character.

\subsubsection{Character-Plugin Obtaining}\label{subsec313}

A Stable Diffusion pipeline with a fine-tuned text-encoder only has the capability to generate an image containing a single character. Moreover, it is impractical to store a separate text-encoder for each individual character. To address the aforementioned concerns, we propose a novel character-plugin obtaining approach that leverages the fine-tuned text-encoder to efficiently generate a lightweight plugin. Furthermore, multiple character-plugins can be seamlessly integrated into an original Stable Diffusion pipeline for story visualization without any interference. The right part of the Fig. \ref{fig2} depicts the flow of the character-plugin obtaining. 

\textbf{\emph{First}}, we organize a token matrix that can be represented as:
\begin{equation}
\bm{TM}=\left[\begin{matrix}bt&ct&et&\ldots&pt&pt\\pt&bt&ct&\ldots&pt&pt\\\vdots&\vdots&\vdots&\ddots&\vdots&\vdots\\pt&pt&pt&\ldots&ct&et\\\end{matrix}\right],\label{eq5}
\end{equation}
where $bt$, $ct$, $et$ and $pt$, respectively, represent the values of \verb|bos_token|, character token, \verb|eos_token| and \verb|pad_token|; $\bm{TM}\in\mathbb{F}^{Q\times{L}}$. The token matrix is designed based on the principle of sequentially sliding the character token ($ct$) from left to right in the token sequence, while simultaneously recording the corresponding token sequences. Each row of the token matrix is formed by a token sequence with different position of character token. In each row of the token matrix, we assign a \verb|bos_token| to the left of the character token, an \verb|eos_token| to the right of the character token, and \verb|pad_tokens| to all other positions. 

\textbf{\emph{Second}}, we input the token matrix into the fine-tuned text-encoder model, resulting in an embedding matrix ($\bm{EM}$):
\begin{equation}
\bm{EM}=E^t(\bm{TM}).\label{eq6}
\end{equation}
The embedding matrix ($\bm{EM}$) has dimensions of $Q\times{L}\times{H}$ and can be expressed as:
\begin{equation}
\begin{aligned}
&\bm{EM}=\\
&\left[\begin{matrix}{\bm{bte}}_*&{\bm{cte}}_{0,1}&{\bm{ete}}_*&\ldots&{\bm{pte}}_*&{\bm{pte}}_*\\{\bm{pte}}_*&{\bm{bte}}_*&{\bm{cte}}_{1,2}&\ldots&{\bm{pte}}_*&{\bm{pte}}_*\\\vdots&\vdots&\vdots&\ddots&\vdots&\vdots\\{\bm{pte}}_*&{\bm{pte}}_*&{\bm{pte}}_*&\ldots&{\bm{cte}}_{Q-1,L-2}&{\bm{ete}}_*\\\end{matrix}\right],\label{eq7}
\end{aligned}
\end{equation}
where $\bm{bte}_{q,l}\in\mathbb{F}^{1\times{H}}$, $\bm{cte}_{q,l}\in\mathbb{F}^{1\times{H}}$, $\bm{ete}_{q,l}\in\mathbb{F}^{1\times{H}}$ and $\bm{pte}_{q,l}\in\mathbb{F}^{1\times{H}}$, respectively, indicate the embedding representations of \verb|bos_token|, character token, \verb|eos_token| and \verb|pad_token| in the $q$th row and $l$th column of the token matrix. In embedding matrix ($\bm{EM}$), there exists a distinct character token embedding ($\bm{cte}$) in each row, and the position of $\bm{cte}$ varies across rows. 

\textbf{\emph{Third}}, we extract the character token embedding from each row of the embedding matrix ($\bm{EM}$) and aggregate them to form a new matrix, referred to as character-plugin ($\bm{CP}$):
\begin{equation}
\bm{CP}=\left[\begin{matrix}{\bm{cte}}_{0,1}&{\bm{cte}}_{1,2}&\ldots&{\bm{cte}}_{Q-1,L-2}\\\end{matrix}\right],\label{eq8}
\end{equation}
where $\bm{cte}_{q,l}$ indicates the character token embedding in the $q$th row and $l$th column of embedding matrix ($\bm{EM}$); $\bm{CP}\in\mathbb{F}^{{(L-2)}\times{H}}$. 

The character-plugin matrix captures the embeddings of character tokens at various positions, enabling seamless integration of these embeddings into the original Stable Diffusion pipeline for story visualization. For large-scale commercial scenarios involving diverse narratives, it is sufficient to store a shared diffusion model along with numerous portable character-plugins.

\subsection{Plugin-Guided and Layout-Guided Inference}\label{subsec32}
In conventional text-to-image inference, the diffusion model demonstrates its capability to generate images based on a given prompt. In the study, we propose an innovative inference strategy that can simultaneously utilize input prompt, character-plugins, and user-defined layout for enhanced story visualization. The implementation of plugin-guided and layout-guided inference involves the token embeddings fusion and cross-attention editing, which will be further elucidated in subsequent sections.

\begin{figure}[ht]
\centering
\includegraphics[width=0.49\textwidth]{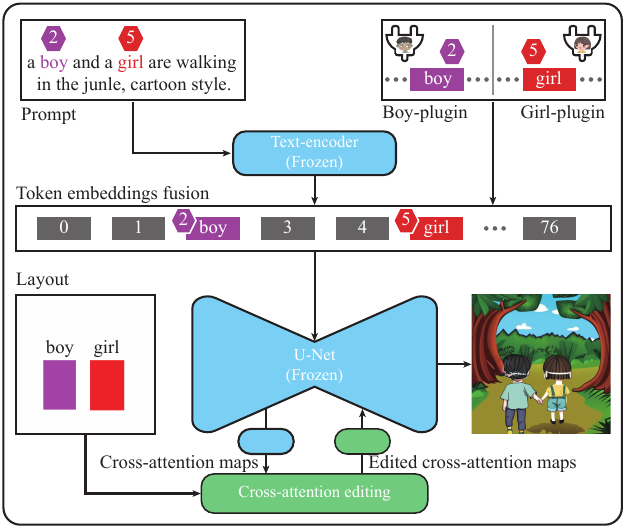}
\caption{Demonstration of the strategy of plugin-guided and layout-guided inference.}\label{fig3}
\end{figure}

\subsubsection{Token Embeddings Fusion for Plugin-Guided Inference}\label{subsec321}

The latest studies related to text-to-image generation demonstrate that manipulation of token embeddings can effectively accomplish image editing \citep{cc45,cc46,cc47,cc48}. We innovatively extend this technique to the domain of story visualization for placing user-defined characters in a generated image with multiple character-plugins. 

\textbf{\emph{First}}, as illustrated in the example depicted in Fig. \ref{fig3}, the prompt is inputted into the text-encoder to derive the token embeddings ($\bm{EB}$), which can be mathematically represented as follows: 
\begin{equation}
\bm{EB}=E^f(\bm{P}),\label{eq9}
\end{equation}
where $\bm{P}$ indicates the prompt tokens; $E^f$ is the text-encoder in original Stable Diffusion pipeline. The $\bm{EB}$, as illustrated in Fig. \ref{fig3}, is a sequence of token embeddings that can be denoted as $\left[\begin{matrix}\ldots&\bm{e}_a^1&\bm{e}_{boy}^2&\bm{e}_{and}^3&\bm{e}_a^4&\bm{e}_{girl}^5&\ldots\\\end{matrix}\right]$.

\textbf{\emph{Second}}, we fuse the token embeddings ($\bm{EB}$) and the multiple character-plugins to obtain the fused token embeddings ($\bm{FEB}$):
\begin{equation}
\bm{FEB}=Fu(\bm{EB},\ \sum_{i=1}^{m}{\bm{CP}}_i).\label{eq10}
\end{equation}
As exemplified in Fig. \ref{fig3}, in the fusion process, we eliminate the embeddings of 'girl' and 'boy' from $\bm{EB}$, followed by fusing the embeddings of 'girl' and 'boy' at specific positions within girl-plugin and boy-plugin into $\bm{EB}$ to construct fused embeddings $\bm{FEB}$. The formulation of $\bm{FEB}$ can be expressed as $\left[\begin{matrix}\ldots&\bm{e}_a^1&\bm{cte}_{*,2}^{boy}&\bm{e}_{and}^3&\bm{e}_a^4&\bm{cte}_{*,5}^{girl}&\ldots\\\end{matrix}\right]$ where $\bm{cte}_{*,2}^{boy}$ and $\bm{cte}_{*,5}^{girl}$ are respectively token embeddings in girl-plugin and boy-plugin. 

The fused token embeddings not only incorporate the contextual information provided by the original text-encoder but also encompass user-defined character information, thereby significantly enhancing the efficacy of story visualization. We then input the fused token embeddings into a U-Net in original Stable Diffusion pipeline.

\subsubsection{Cross-Attention Editing for Layout-Guided Inference}\label{subsec322}

Users often have a strong inclination to regulate the layout of the produced images. However, the diffusion model's output exhibits inherent randomness. Previous studies \citep{cc18,cc19,cc45,cc49} have demonstrated the feasibility of manipulating the layout of generated images through direct manipulation of cross-attention maps in U-net. We also employ this technique in the domain of narrative visualization. Similar to Cones 2 \citep{cc19}, we employ a predetermined layout image and integrate it with the corresponding cross-attention map for each character, thereby exerting precise control over the spatial locations of characters in the generated results. The lower part of Fig. \ref{fig3} illustrates the process of cross-attention editing.  

\textbf{\emph{First}}, we extract the character-related cross-attention maps in the U-Net. The specific cross-attention maps corresponding to each character can be denoted as ${\bm{CAM}}_c|c=["girl","boy",\ldots]$.

\textbf{\emph{Second}}, we divide a single layout image with multiple character position boxes into multiple layout images, each containing only one character position box, denoted as ${\bm{LO}}_c|c=["girl","boy",\ldots]$. The position box region in each layout image is assigned a positive numerical value, while the remaining region is assigned a negative numerical value. We directly add the layout images to the cross-attention maps for the purpose of obtaining the edited maps:
\begin{equation}
{\bm{ECAM}}_c={\bm{CAM}}_c\oplus{[\xi(t)\bullet\bm{LO}}_c],\label{eq11}
\end{equation}
where $c=["girl","boy",\ldots]$; the symbol of $\oplus$ represents the operation of matrix addition; the function $\xi(\bullet)$ exhibits a decaying trend as the number of inference step $t$ increases, serving as a means to regulate the intensity of editing.

\textbf{\emph{Thrid}}, we input the edited cross-attention maps into the U-Net for final inference. The edited cross-attention maps incorporates user-specified positional information for each character, enabling the U-Net to produce images that conform to the specified layout provided by the user.

\section{Experiments}\label{sec4}

\subsection{Character Dataset Preparation}\label{subsec41}

\begin{figure*}[ht]%
\centering
\includegraphics[width=0.99\textwidth]{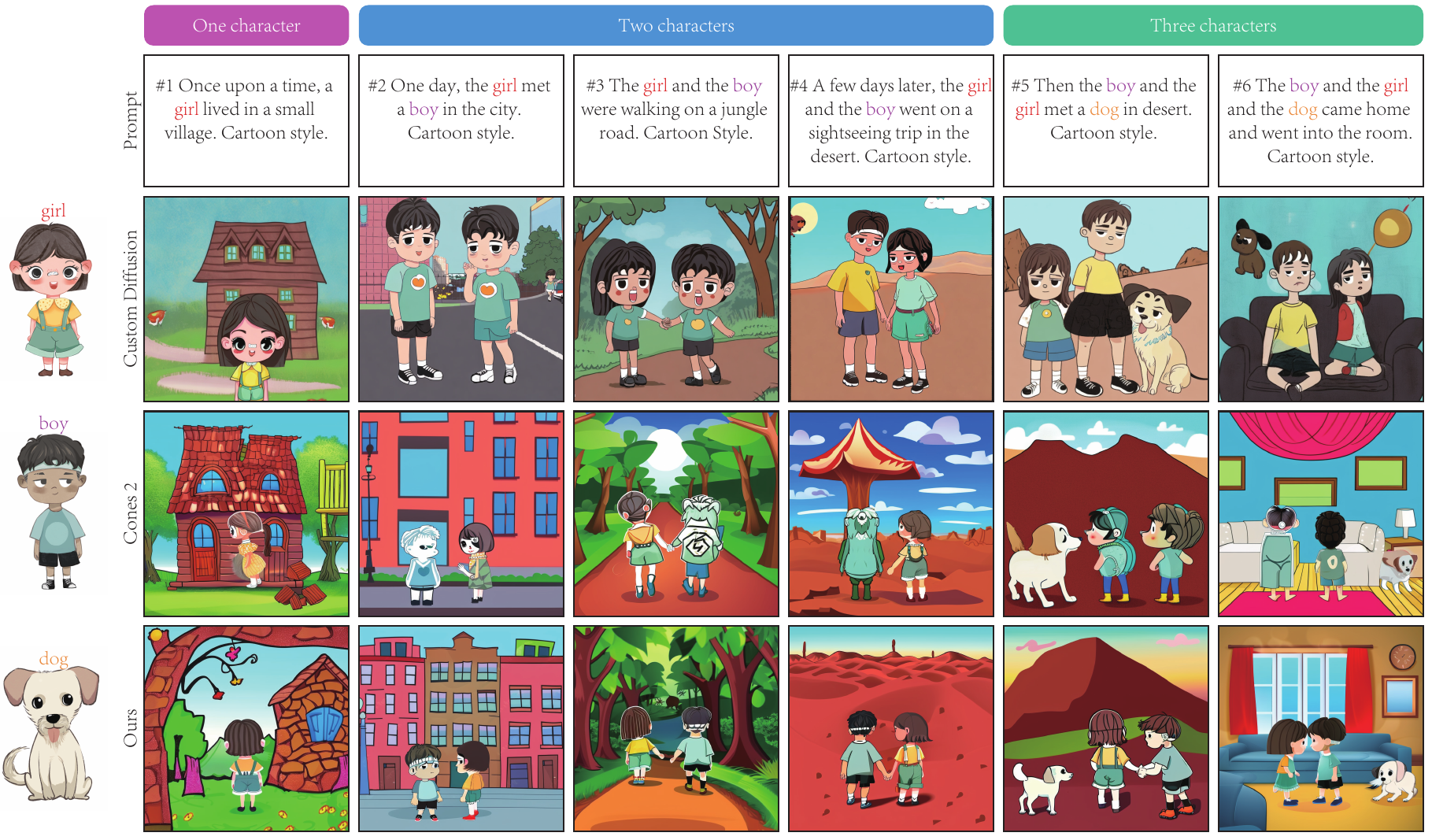}
\caption{\textbf{Qualitative comparison} of story visualization ability between our method and baselines. We use three characters - a girl, a boy, and a dog - to visualize a story with 7 prompts: one prompt features only one character, three prompts involve two characters, and the remaining two prompts include all three characters.}\label{fig4}
\end{figure*}

The process of storybook creation typically involves two sequential steps: firstly, conceptualizing the visual representations of characters; secondly, visualizing a narrative based on the conceived characters and storyline. The first step involves generating a limited number of examples, specifically where each character is represented by a small set of exemplary images. The second step focuses on producing an extensive range of samples that depict the unfolding storyline. The first process is brief and does not require significant human resources, while the second process is time-consuming and demands substantial human resources. Previous methods, such as StoryDALL-E \citep{cc11} and Make-A-Story \citep{cc12}, heavily rely on extensive data, rendering them effective only after accumulating a substantial number of samples during the second stage of storybook creation. Consequently, the efficacy of these conventional algorithms is significantly diminished.

Based on the aforementioned considerations, we have developed a character dataset to validate the effectiveness of CogCartoon in creating storybooks efficiently: during the initial phase, a designer creates a limited number of character images; subsequently, our method is employed to directly visualize the narrative. We hired a professional character designer, who took 24 working days to design three characters - a girl, a boy, and a dog - with 40 samples for each character.

\subsection{Settings}\label{subsec42}

\subsubsection{Evaluation metrics}\label{subsec421}

We evaluate CogCartoon with two settings: (1) quantitative assessment with text alignment (TA) and image alignment (IA) \citep{cc39,cc19}; (2) human evaluation with correspondence (COR), coherence (COH) and quality (QUA) \citep{cc20}. The TA has the capability to evaluate the semantic relevance between the input prompts and produced images. The IA is capable of measuring the visual similarity between the target characters and the produced images. For visualizing stories with multiple characters, we compute the visual similarity between each target character and produced images individually, and subsequently calculate their average value. The assessment of COR, COH and QUA is predicated upon the subsequent inquiries. (1) Does the produced image precisely represent the input prompt? (2) Is the character in the produced image consistent with the target character? (3) What is the quality of the produced image? The score for human evaluation ranges from 0 to 3, with higher score indicating superior generation quality. The FID metric \citep{cc50} is not suitable for evaluation due to its reliance on a substantial number of benchmark samples.

\subsubsection{Baselines}\label{subsec422}

As mentioned earlier, the conventional methods for story visualization are not applicable to scenarios with limited training samples. The customized image generation methods offer a promising solution to visualize narratives, even when the training dataset is limited. Therefore, we compare CogCartoon with two state-of-the-art multi-subject customization technologies, namely Custom Diffusion \citep{cc18} and Cones 2 \citep{cc19}. We exclude DreamBooth \citep{cc16}, Painting-by-Example \citep{cc38} and Textual Inversion \citep{cc39} techniques due to their suitability for single-subject customization tasks, whereas story visualization often involves multiple characters. 

\subsubsection{Implementation details}\label{subsec423}

To guarantee a fair comparison, we employ the Stable Diffusion v2.1 \citep{cc10} as the pre-trained model for all methods. During inference, we utilize 100 steps of DDIM \citep{cc53} and set the scale parameter to 7.5. The experiments are conducted using a workstation equipped with one A-100 GPU.

\noindent\textbf{Custom Diffusion.} We implement Custom Diffusion with the third-party toolkit called Diffusers \citep{cc53}. During training, we set the prior loss weight to 1.0, the learning rate to 1e-5, the batch size to 1 and the training step to 5000. We have observed that achieving satisfactory results is challenging with a few hundred training steps, whereas several thousand training steps can yield decent outcomes.

\noindent\textbf{Cones 2.} We implement Cones 2 with the official code. For the training, we employ the default parameter settings specified in the code, which encompass a learning rate of 5e-6 and a batch size of 1. We do not set the training step to the default value of 4000 because the generated results are very poor. Through extensive experimentation, we have found that setting the training step to 400k results in acceptable generation. During the inference process of the Cones 2, a positive parameter and a negative parameter are utilized to restrict the object's position in the generated image. The values for these parameters are set to +2.5 and -1e8, respectively.

\noindent\textbf{Ours.} In character-plugin generation, we set the number of augmented images as 300. We train the text-encoder with a training step of 400k, a learning rate of 5e-6 and a batch size of 1. During inference, similar to Cones 2, we utilize a positive parameter of +2.5 and a negative parameter of -1e-8 to control the character's position in the generated image.

% Table generated by Excel2LaTeX from sheet
\begin{table*}[h]
  \caption{\textbf{Quantitative comparisons} with text alignment (TA) and image alignment (IA). Higher score denotes superior performance.}
    \begin{tabular*}{\textwidth}{@{\extracolsep\fill}crccccccc} %\begin{tabular}{crccccccc}
    \toprule
          &       & \multicolumn{3}{c}{One Character} & \multicolumn{3}{c}{Two Characters} & \multicolumn{1}{c}{Three Characters} \\
\cmidrule{3-5}\cmidrule{6-8}\cmidrule{9-9}          &       & story 1 & story 2 & story 3 & \multicolumn{1}{c}{story 4} & \multicolumn{1}{c}{story 5} & \multicolumn{1}{c}{story 6} & \multicolumn{1}{c}{story 7} \\
          &       & \multicolumn{1}{c}{girl} & \multicolumn{1}{c}{boy} & \multicolumn{1}{c}{dog} & \multicolumn{1}{c}{girl\&boy} & \multicolumn{1}{c}{girl\&dog} & \multicolumn{1}{c}{boy\&dog} & \multicolumn{1}{c}{girl\&boy\&dog} \\
    \midrule
    \multirow{1.5}[2]{*}{Cones2} & TA &0.320       &\textbf{0.319}       &0.318       &0.322       &0.338       &\textbf{0.340}       &0.337  \\
          & IA &0.695       &0.697       &0.746       &0.661       &0.659       &0.657       &0.656  \\
    \midrule
    \multirow{1.5}[2]{*}{Ours} & TA &\textbf{0.326}       &0.317       &\textbf{0.321}       &\textbf{0.339}       &\textbf{0.339}       &0.335       &\textbf{0.340}  \\
          & IA &\textbf{0.722}       &\textbf{0.722}       &\textbf{0.747}       &\textbf{0.681}       &\textbf{0.669}       &\textbf{0.667}       &\textbf{0.659}  \\
    \bottomrule
    \end{tabular*}%
  \label{table1}%
\end{table*}%

% Table generated by Excel2LaTeX from sheet
\begin{table*}[h]
  \caption{\textbf{Human evaluation} on correspondence (COR), coherence (COH) and quality (QUA). Higher score denotes superior performance.}
    \begin{tabular*}{\textwidth}{@{\extracolsep\fill}crccccccc} %\begin{tabular}{crccccccc}
    \toprule
          &       & \multicolumn{3}{c}{One Character} & \multicolumn{3}{c}{Two Characters} & \multicolumn{1}{c}{Three Characters} \\
\cmidrule{3-5}\cmidrule{6-8}\cmidrule{9-9}          &       & story 1 & story 2 & story 3 & \multicolumn{1}{c}{story 4} & \multicolumn{1}{c}{story 5} & \multicolumn{1}{c}{story 6} & \multicolumn{1}{c}{story 7} \\
          &       & \multicolumn{1}{c}{girl} & \multicolumn{1}{c}{boy} & \multicolumn{1}{c}{dog} & \multicolumn{1}{c}{girl\&boy} & \multicolumn{1}{c}{girl\&dog} & \multicolumn{1}{c}{boy\&dog} & \multicolumn{1}{c}{girl\&boy\&dog} \\
    \midrule
    \multirow{2.5}[2]{*}{Cones2} 
          & COR &1.31       &1.59       &1.77       &0.27       &0.94       &1.37       &0.53  \\
          & COH &0.83       &0.71       &1.50       &0.23       &0.99       &1.36       &0.54  \\
          & QUA &0.76       &0.72       &1.30       &0.10       &1.00       &1.33       &0.56  \\
    \midrule
    \multirow{2.5}[2]{*}{Ours}
          & COR &\textbf{2.87}       &\textbf{2.93}       &\textbf{2.91}       &\textbf{2.33}       &\textbf{2.43}       &\textbf{2.66}       &\textbf{1.73}  \\
          & COH &\textbf{2.85}       &\textbf{2.79}       &\textbf{2.77}       &\textbf{2.40}       &\textbf{2.35}       &\textbf{2.51}       &\textbf{1.76}  \\
          & QUA &\textbf{2.79}       &\textbf{2.80}       &\textbf{2.75}       &\textbf{2.32}       &\textbf{2.12}       &\textbf{2.40}       &\textbf{1.69}  \\
    \bottomrule
    \end{tabular*}%
  \label{table2}%
\end{table*}%

\subsection{Main Results}\label{subsec43}

\subsubsection{Qualitative Comparison}\label{subsec431}

\textbf{CogCartoon possesses stronger capabilities in visualizing stories compared to baselines.} As depicted in Fig. \ref{fig4}, CogCartoon has the ability to visually depict narratives encompassing expansive scenarios, such as village, city, jungle, desert, and room. It is capable of generating high-quality illustrations featuring multiple characters based on given prompts, ensuring a remarkable level of consistency between the generated characters and target characters. The other two baseline methods, however, struggle to generate satisfactory images based on the given prompts. Specifically, the utilization of Custom Diffusion in the creation of stories involving two characters leads to a significant attribute confusion. For example, in Figure \ref{fig4}, the second column of the first row shows two boys instead of one girl and one boy as expected. Despite the absence of attribute confusion, Cones 2 fails to adequately represent the appearances of the characters, leading to a limited consistency between the produced images' characters and the target characters.

\begin{figure}[ht]%
\centering
\includegraphics[width=0.49\textwidth]{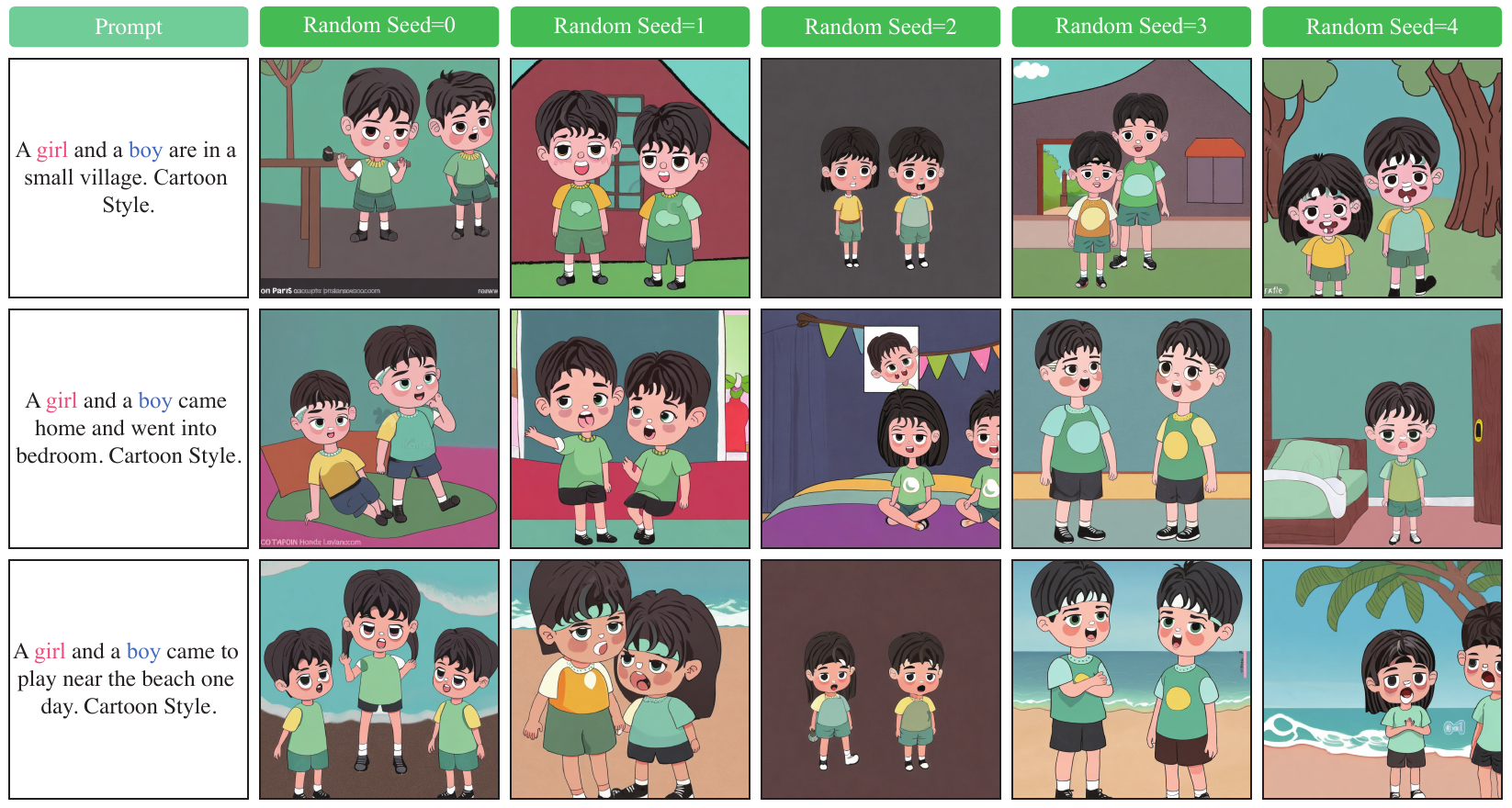}
\caption{Further experimental results on Custom Diffusion with two characters.We select three distinct prompts, each comprising five unique random seeds, resulting in a total of 15 images.}\label{fig5}
\end{figure}

\begin{figure*}[ht]%
\centering
\includegraphics[width=0.99\textwidth]{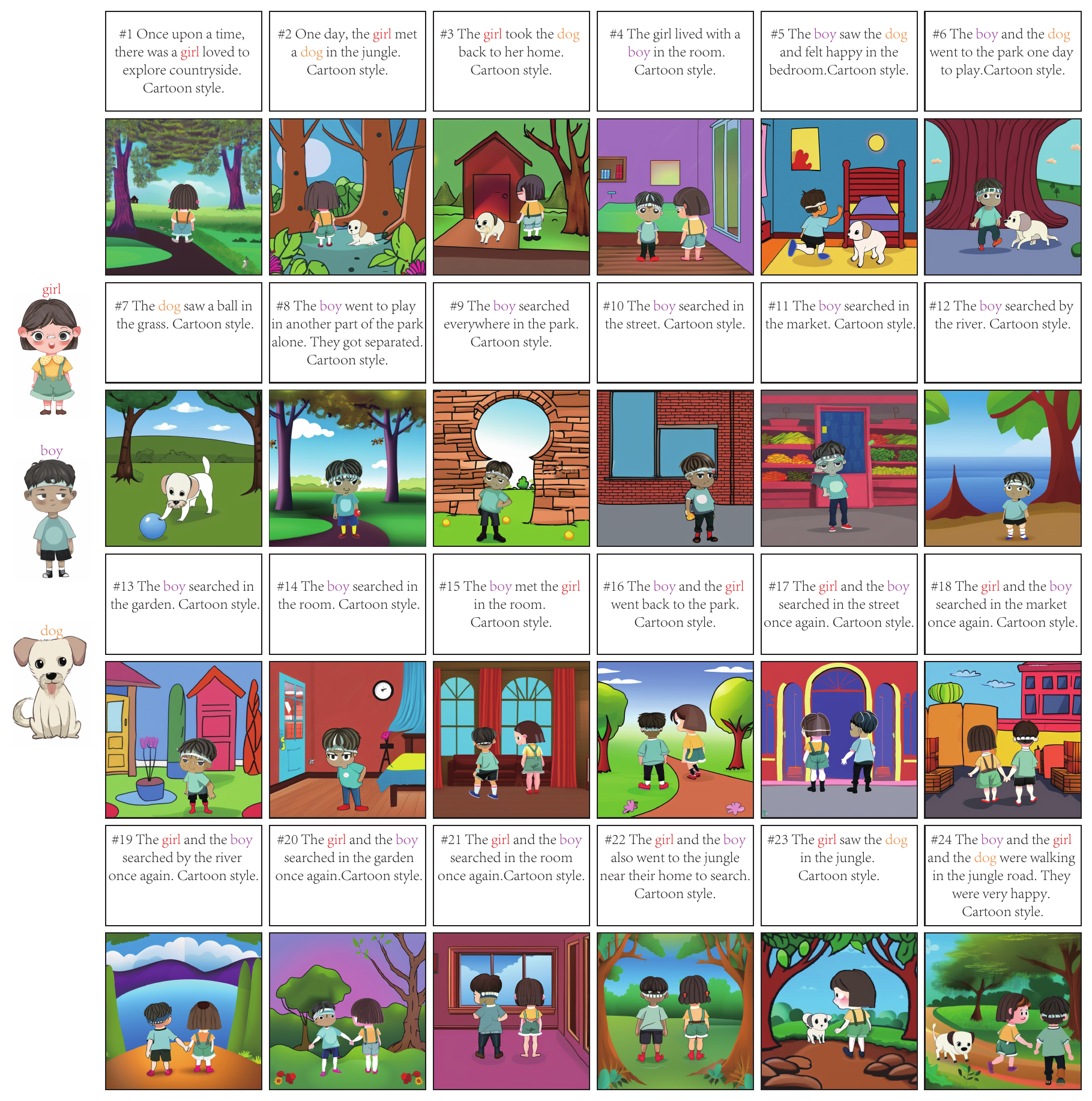}
\caption{Long story visualization results achieved through the CogCartoon. We use CogCartoon to visualize a long story with 24 frames that span across a jungle, a room, a park, etc.}\label{fig6}
\end{figure*}

\begin{figure*}[ht]%
\centering
\includegraphics[width=0.99\textwidth]{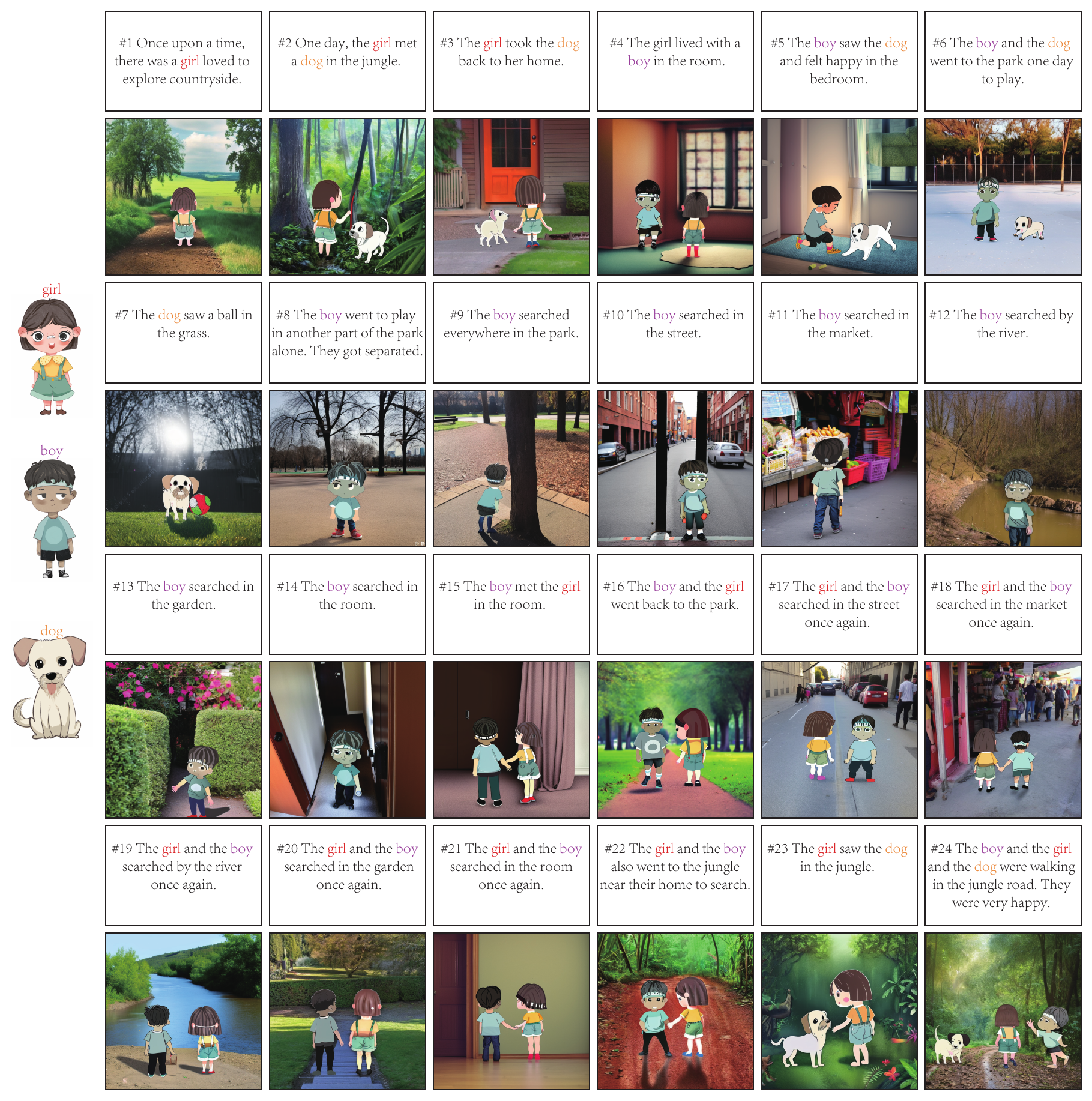}
\caption{Realistic style story visualization results achieved through the CogCartoon.We use CogCartoon to visualize a long, realistic style story with 24 frames across multiple scenes. Realistic style story pictures can be generated using CogCartoon by simply removing the text description "Cartoon Style" from the prompt.}\label{fig7}
\end{figure*}

Prior to delving into the quantitative comparison experiments, it is necessary to provide a comprehensive elucidation of the findings depicted in Fig. \ref{fig5}. The results of additional experiments on Custom Diffusion for visualizing the narrative of two characters are presented in Fig. \ref{fig5}, considering different prompts and random seeds. We have observed that Custom Diffusion, when applied in diverse settings, is prone to a significant attribute confusion. In fact, the Custom Diffusion study also highlight a similar phenomenon in the generation of pet cats and dogs \citep{cc18}. The conclusions we obtained are consistent with those derived from the Custom diffusion experiments. Additionally, Custom Diffusion necessitates retraining all the data upon the inclusion of new characters. Based on the aforementioned experimental results and findings, employing Custom Diffusion as a baseline for subsequent quantitative comparisons and manual evaluations is deemed unnecessary.

\subsubsection{Quantitative Comparison}\label{subsec432}

\textbf{CogCartoon achieves higher metrics compared to the baseline.} We designed seven stories comprising cases featuring various character combinations: three of them featured a single character, three of them featured two characters, and one of them featured three characters. Each story comprises five prompts, with each prompt consisting of 20 randomly generated seeds, resulting in a total of 700 images for the quantitative comparison experiment. CogCartoon achieves better results than the baseline method in almost all settings, as evidenced by its higher IA and TA scores (Table \ref{table1}). This indicates that the generated images by CogCartoon effectively capture the content of the prompt and exhibit character consistency. Despite the remarkable achievements of Cones 2 in the realm of multi-subject customization, its application to story visualization poses significant challenges.

\subsubsection{Human Evaluation}\label{subsec433}

\textbf{CogCartoon is more preferred by human than the baseline.} The above 700 images were manually evaluated in terms of correspondence, coherence, and quality. As presented in Table \ref{table2}, CogCartoon consistently outperforms other approach across all metrics and experimental settings, thereby indicating its ability to generate images that align more closely with human preferences.

\subsection{Toward Challenging Tasks}\label{subsec44}

\begin{figure}[ht]%
\centering
\includegraphics[width=0.49\textwidth]{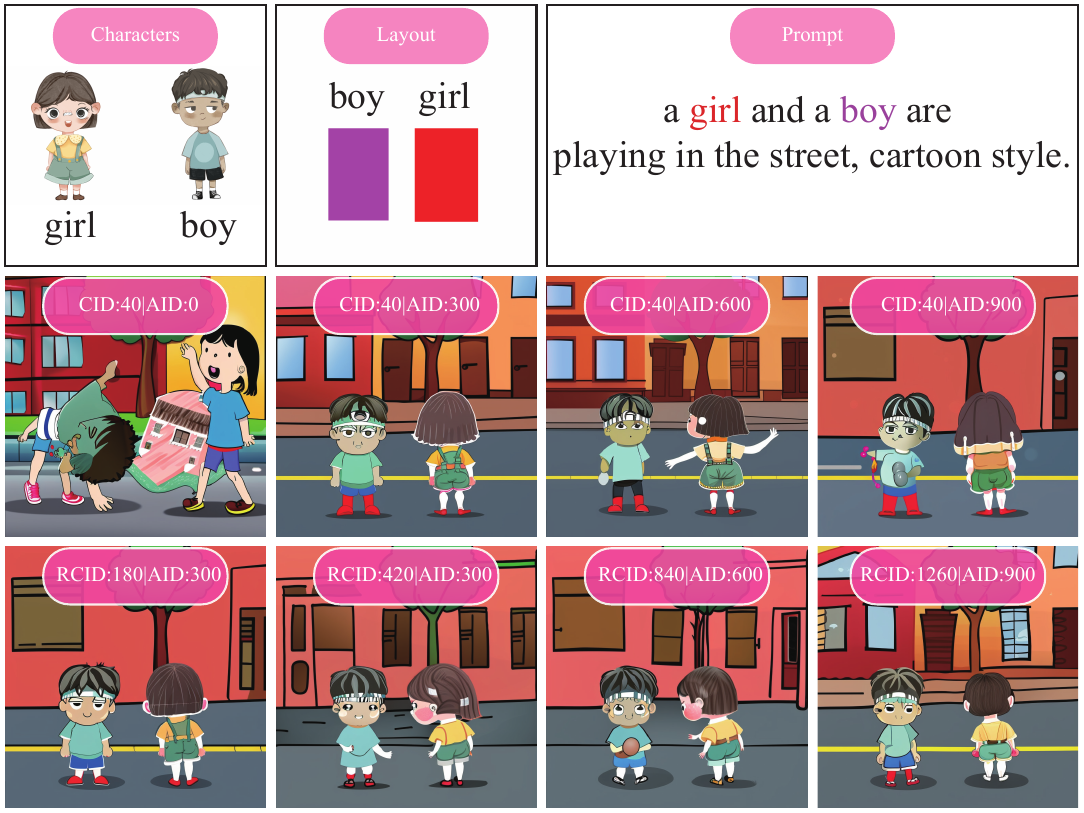}
\caption{\textbf{Ablation study} on the number of augmented images. We maintain fixed factors including characters, layout, prompt, and random seed while utilizing CogCartoon to generate story images with varying amounts of augmented data. CID indicates the number of images in the character image dataset; AID indicates the number of images in the augmented image dataset; RCID indicates the number of images in the dataset formed by repeatedly copying the character dataset.}\label{fig8}
\end{figure}

\begin{figure}[ht]%
\centering
\includegraphics[width=0.49\textwidth]{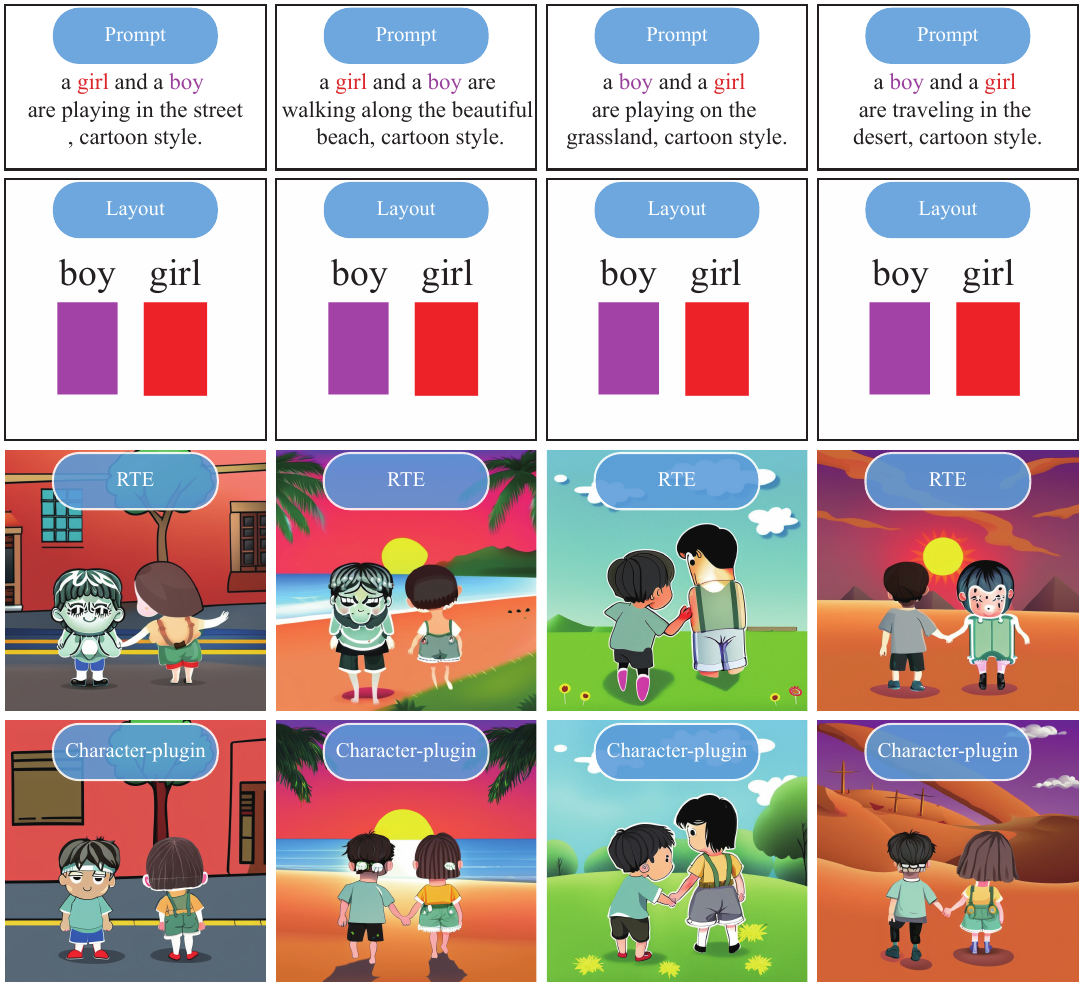}
\caption{\textbf{Ablation study} on plugin characterization. We fix the characters, layout and random seed, and use CogCartoon to generate story images with different prompts. For each prompt, we select the Residual Token Embedding (RTE) and our proposed character-plugin for comparison.}\label{fig9}
\end{figure}

\textbf{CogCartoon demonstrates strong capabilities in handling more challenging tasks, including long story visualization and realistic style story visualization.} To further exemplify the advanced capabilities of CogCartoon, we subject it to more challenging tasks. The task of visualizing a long story is to generate coherent images that cover a large number of scenes from a lengthy narrative, typically consisting of more than dozens of sentences. As depicted in Fig. \ref{fig6}, CogCartoon is capable of generating story images in massive scenes due to the retention of all weights from the original Stable Diffusion model, enabling visualization of long stories. Traditional methods of visualizing stories, such as StoryDall-E \citep{cc11} and Make-A-Story \citep{cc12}, require fine-tuning the weights of the base model using a specific dataset. This limitation results in a restricted variety of scenes generated by these methods, making them unsuitable for long story visualization tasks.

The task of realistic style story visualization aims to generate story images based on given prompts, where the background exhibits real-world characteristics while the characters are depicted in a cartoon style. Realistic style cartoon storybooks are popular among anime enthusiasts. The task of visualizing realistic style stories necessitates the model's ability to concurrently generate a story image in two distinct styles. As shown in Fig. \ref{fig7}, thanks to the proposed character-plugin generation and inference methods, CogCartoon can generate cartoon characters flawlessly within realistic backgrounds, thereby successfully accomplishing challenging task of realistic style story visualization. Traditional story visualization methods \citep{cc11,cc12,cc13} can only generate stories with a single style, making it difficult for them to accomplish the tasks of visualizing stories in a realistic style.

\subsection{Ablation Studies}\label{subsec45}

\subsubsection{Ablation study on the number of augmented images}\label{subsec451}

\textbf{CogCartoon demonstrates robustness to the number of augmented data.} The data augmentation operation is an important part of CogCartoon, which we are exploring in greater detail. Fig. \ref{fig8} displays the experimental results with varying amounts of augmented data, leading to two conclusions: 1. Augmentation proves to be an extremely effective method for enhancing story visualization (comparing the image in the second row of the first column of Fig. \ref{fig8} with the remaining images). 2. Regardless of changes in the number of augmented samples, a consistent and satisfactory visual representation of the story is achieved.

\subsubsection{Ablation study on plugin characterization}\label{subsec452}

\textbf{CogCartoon is capable of accurately representing characters.} Characterizing characters using the character-plugin is an important aspect of CogCartoon, which we will further investigate. The use of Residual Token Embedding (RTE) has demonstrated its efficacy in subject representation. The RTE can be seamlessly integrated into a pretrained diffusion model to generate an image featuring a specific subject \citep{cc51,cc19}. We compared the results of story visualization using RTE and our proposed character-plugin with different prompts but the same random seed and layout. The experimental outcomes demonstrate that the character-plugin proposed in CogCartoon effectively portrays the intricacies of the characters and presents them in the generated images (Fig. \ref{fig9}). Although RTE is effective in representing subjects, it cannot be used to represent characters.

\section{Conclusion}\label{sec5}

We present CogCartoon, a practical approach for story visualization. By incorporating the modules of character-plugin generation and plugin-guided and layout-guided Inference, CogCartoon enables story visualization even with limited training samples and storage resources, while also allowing the user to control the layout and insert new characters. We conduct a multitude of experiments, encompassing qualitative analysis, quantitative analysis, and human evaluation, which unequivocally demonstrate the exceptional performance of our method in story visualization. Furthermore, CogCartoon demonstrates its capability in accomplishing challenging tasks, including long story visualization and realistic style story visualization. Conventional methods for visualizing stories prove inadequate and customized image generation techniques fail to yield satisfactory outcomes when dealing with only a limited number of training samples. Consequently, the significance of CogCartoon in the realm of story visualization is heightened.

%%===========================================================================================%%
\backmatter

\bmhead*{Acknowledgments}

 We express our gratitude to Xiaoting Lu for her contribution in producing the character dataset, and extend our appreciation to Anqi Liu for her valuable insights and discussions. 

%%\bibliography{sn-bibliography}% common bib file

\end{document}